\documentclass{article} %
\usepackage{iclr2026_conference,times}

\usepackage{amsmath,amsfonts,bm}

\def\eqref#1{equation~\ref{#1}}

\def\1{\bm{1}}

\DeclareMathAlphabet{\mathsfit}{\encodingdefault}{\sfdefault}{m}{sl}
\SetMathAlphabet{\mathsfit}{bold}{\encodingdefault}{\sfdefault}{bx}{n}

\usepackage{microtype}
\usepackage{graphicx}
\usepackage{subcaption}
\usepackage{booktabs} %

\usepackage{microtype}
\usepackage{graphicx}
\usepackage{booktabs} %
\usepackage{enumitem}
\usepackage{comment}
\usepackage{makecell}
\setlist[itemize]{itemsep=0em} %
\setlist[enumerate]{itemsep=0em}

\usepackage{amsmath, amssymb, amsthm}
\usepackage{bm}                  %
\usepackage{mathtools}           %
\usepackage{booktabs}            %
\usepackage{multirow}            %
\usepackage{array}               %
\usepackage{enumitem}            %

\usepackage{graphicx}
\usepackage{subcaption}          %
\usepackage{float}               %
\usepackage{adjustbox}           %
\usepackage{colortbl, xcolor}    %

\usepackage{hyperref}
\usepackage{xcolor}         %

\newcommand{\code}[2][]{\mathsf{\mathtt{#2}_{\text{#1}}}}

\usepackage{algorithm}
\usepackage[noend]{algpseudocode}       %
\usepackage{algorithmicx}        %
\usepackage[skip=0pt]{caption}             %
\usepackage{listings}            %

\definecolor{promptbg}{RGB}{240,240,240}
\definecolor{promptframe}{RGB}{204,204,204}

\lstdefinestyle{promptstyle}{
    backgroundcolor=\color{promptbg},
    frame=single,
    framesep=3pt,
    rulecolor=\color{promptframe},
    basicstyle=\ttfamily\small,
    breaklines=true,
    breakatwhitespace=true,
    showstringspaces=false,
    columns=flexible,
    keepspaces=true,
    numbers=none,
    commentstyle=\color{gray},
    keywordstyle=\color{black},
    stringstyle=\color{black},
    captionpos=b
}

\usepackage{tcolorbox}
\usepackage{listings}
\usepackage{xcolor} %
\tcbuselibrary{listings,breakable,xparse,skins}

\definecolor{bg}{gray}{0.95}
\definecolor{codekw}{RGB}{33,74,135}      %
\definecolor{codestring}{RGB}{124,34,93}  %
\definecolor{codecomment}{RGB}{0,110,0}   %

\lstdefinestyle{code}{
  basicstyle=\ttfamily\small,
  keywordstyle=\color{codekw}\bfseries,
  stringstyle=\color{codestring},
  commentstyle=\color{codecomment},
  showstringspaces=false,
  showspaces=false,
  showtabs=false,
  breaklines=true,
  columns=fullflexible,
}

\DeclareTCBListing{codebox}{O{}m!O{}}{%
  breakable=true,
  listing engine=listings,
  listing only,
  listing options={%
    style=code,
    language=#2,
    numbers=left,
    numbersep=8pt,
    #1},
  boxsep=0pt,
  left skip=0pt,
  right skip=0pt,
  left=25pt,
  right=0pt,
  top=3pt,
  bottom=3pt,
  arc=5pt,
  leftrule=0pt,
  rightrule=0pt,
  bottomrule=2pt,
  toprule=2pt,
  colback=bg,
  colframe=orange!70,
  enhanced,
  overlay={%
    \begin{tcbclipinterior}
      \fill[orange!20!white] (frame.south west) rectangle ([xshift=20pt]frame.north west);
    \end{tcbclipinterior}
  },
  #3}

\usepackage{hyperref}
\usepackage{url}

\usepackage{amsmath}
\usepackage{amssymb}
\usepackage{mathtools}
\usepackage{amsthm}

\usepackage[capitalize,noabbrev]{cleveref}

\theoremstyle{plain}

\theoremstyle{definition}

\theoremstyle{remark}

\usepackage[textsize=tiny]{todonotes}

\title{Inference-Time Distillation: Cost-Efficient Agents Without Fine-Tuning or Manual Prompt Engineering}

\author{Vishnu Sarukkai\thanks{Stanford University}, Asanshay Gupta\textsuperscript{*}, James Hong\thanks{Reve}, Michaël Gharbi\textsuperscript{†}, Kayvon Fatahalian\textsuperscript{*}
}

\setlist{nosep}

\iclrfinalcopy %
\begin{document}

\maketitle

Deploying LLM agents at scale typically requires choosing between quality and cost. Existing cost-reduction approaches fail to preserve \emph{agility}: the ability to iterate rapidly without human time bottlenecks. Prompt engineering is brittle and slows iteration, while fine-tuning requires multi-day training and commitment to fixed designs; both are impractical for iterative workflows and time-sensitive batch jobs.
We demonstrate that established inference-time techniques—dynamic in-context learning and self-consistency cascades—can be leveraged to shift the cost-accuracy Pareto frontier while preserving agility. Practitioners run the teacher on a small task subset to collect demonstrations, then immediately deploy a cheaper student on the remainder. At each step, the system retrieves relevant teacher demonstrations as in-context examples. When multiple student samples agree, we proceed; when they diverge, we fall back to the teacher. This requires no prompt engineering or training.
On ALFWorld, we match teacher accuracy at 2.5$\times$ lower cost (\$0.059 → \$0.024 per episode). On AppWorld, we achieve 3.5$\times$ cost reduction while recovering 79\% of teacher accuracy. Our empirical analyses provide guidance on key design choices: teacher database size, demonstration set size, retrieval strategy, and cascade thresholds. These analyses highlight inference-time levers for navigating cost-performance tradeoffs without sacrificing human development speed.

\section{Introduction}
The world has an abundance of ideas for agentic solutions to enable new software features or automate tasks. However, practitioners face a fundamental tension: developing agents requires \emph{agility}—the ability to iterate rapidly without human time bottlenecks—but deploying them at scale requires cost efficiency. Traditional approaches force an unacceptable tradeoff. Prompt engineering can improve model performance, but requires laborious trial-and-error that is brittle, unreliable, and slows iteration cycles~\citep{khattab2023dspy}. Fine-tuning reliably improves the cost-performance frontier, but has high infrastructure overhead and can require days of training, making it impractical when jobs need immediate execution.

We define \emph{agility} as minimizing human development time: the ability to iterate on agent designs and deploy at scale without manual prompt engineering, multi-day training cycles, or trial-and-error tuning. This is critical in two scenarios: (1) \emph{iterative development workflows}, where organizations refine agent architectures based on user feedback while serving production traffic; and (2) \emph{time-sensitive batch jobs}, such as processing large-scale datasets, where running a teacher model on a small subset to bootstrap a cheaper student for the remaining tasks must happen immediately, without waiting for fine-tuning cycles or iterative prompt refinement. 

We demonstrate that the established inference-time mechanisms of dynamic in-context learning~\citep{liu2021makes,zhou2024trad,sarukkai:2025:selfgenerated} and self-consistency cascades~\citep{wang2022selfconsistency,yue2024mot} can be combined to achieve distillation-like cost reductions while preserving agility. The key is \emph{how} these techniques are applied. When running a large batch of tasks, practitioners run the teacher model on a small representative subset (hundreds of episodes) to collect demonstrations, then immediately deploy a cheaper student on the remainder. At each step, the system retrieves relevant teacher demonstrations and inserts them into the student's context. When retrieved examples enable confident student predictions (detected via self-consistency across multiple samples), the student proceeds; otherwise, the system falls back to the teacher. This requires no prompt engineering, model training, or manual tuning.

Across ALFWorld~\citep{shridhar2020alfworld} and AppWorld~\citep{appworld-acl24} benchmarks, our approach shifts the cost-accuracy Pareto frontier. On ALFWorld, we achieve 2.5$\times$ cost reduction while matching teacher-level accuracy (\$0.059 → \$0.024 per episode). On AppWorld, a more complex benchmark requiring multi-step API workflows, we achieve 3.5$\times$ cost reduction while recovering 79\% of teacher accuracy. Crucially, our empirical analyses provide practitioners with guidance on key design choices: teacher database size, demonstration set size, demonstration content, retrieval strategy, and cascade thresholds. These levers enable practitioners to reliably navigate cost-performance tradeoffs without sacrificing human development speed.

\begin{figure*}[t]
    \centering
    \includegraphics[width=1.0\textwidth]{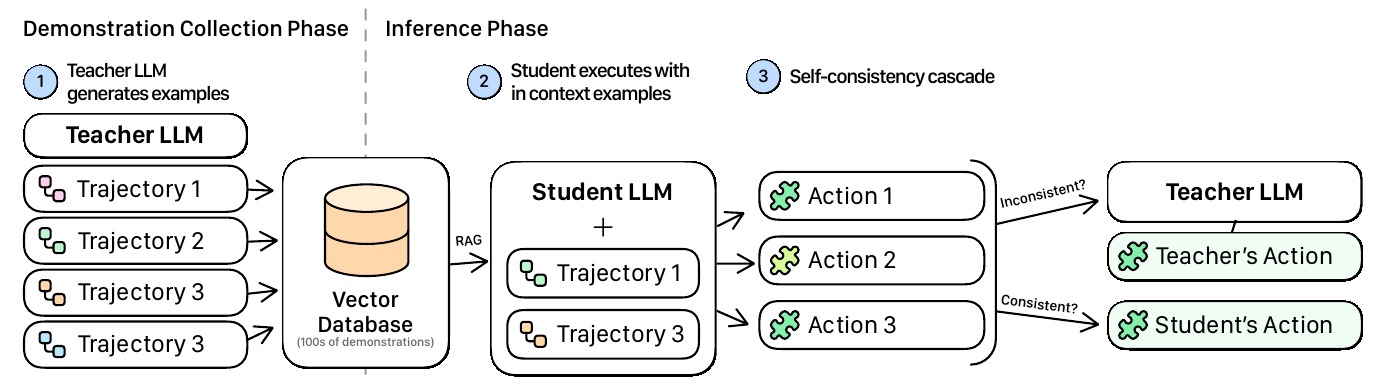}
    \caption{
    \textbf{Our approach combines dynamic in-context learning with self-consistency cascades.} Teacher demonstrations are collected once and indexed. At deployment, relevant examples are retrieved at each step to guide a cheaper student model. Self-consistency across multiple student samples determines when to proceed vs. defer to the teacher.
    }
    \label{fig:method}
    \vspace{-1em}
\end{figure*}

\section{Problem Statement}
\label{sec:problem}

We address cost-effective agent deployment in settings requiring rapid iteration and immediate execution. We focus on multi-step interactive tasks where an agent must take multiple actions, each time receiving feedback from its environment to accomplish a goal (e.g., API call sequences, structured data workflows).

Formally, consider an interactive environment $\mathcal{E}$ in which an agent receives a task specification $(e, g)$ (consisting of an environment instance $e$ and task goal $g$), observes states $o_t$, and executes actions $a_t$ over time steps $t = 1, \dots, T$. Each episode produces a trajectory $\tau = \{(o_t, a_t)\}_{t=1}^{T}$, terminating in success or failure.

We assume access to a high-capacity \emph{teacher} $M_{\text{t}}$ with per-token cost $c_{\text{t}}$ and a lower-cost \emph{student} $M_{\text{s}}$ with cost $c_{\text{s}} \ll c_{\text{t}}$. The teacher is generally more capable. We can collect teacher demonstrations offline from tasks $\mathcal{T}_{\text{demo}}$ and have inference access to both models.

\noindent\textbf{Constraints.} We operate under constraints ruling out traditional cost-reduction approaches: (1) no model training—the policy $\pi$ uses only inference-time mechanisms (retrieval, prompting, routing), and (2) immediate deployment—the system must be ready after demonstration collection, without training cycles. These constraints arise in iterative development workflows (where quick redeployment as agent designs evolve is essential) and time-sensitive batch jobs.

\noindent\textbf{Objective.} We seek a policy maximizing success on test tasks $\mathcal{T}_{\text{test}}$ (with $|\mathcal{T}_{\text{test}}| \gg |\mathcal{T}_{\text{demo}}|$) while minimizing cost:
\[
\pi^* = \arg\max_{\pi \in \Pi_{\text{frozen}}} \mathbb{E}_{(e,g)} \left[
S(\pi, e, g) - \lambda \cdot C(\pi, e, g)
\right]
\]
where $\Pi_{\text{frozen}}$ denotes policies using only frozen models, $\text{S}(\pi, e, g) \in \{0,1\}$ indicates success, $\text{C}(\pi, e, g)$ is total inference cost, and $\lambda > 0$ controls the tradeoff.

The challenge: the student alone has low cost but poor accuracy; the teacher has high accuracy but prohibitive cost at scale. Traditional distillation via fine-tuning violates our constraints (requires training, takes days, breaks when architectures change). We must improve student capability using only inference-time mechanisms.

\section{Methods}
\label{sec:methods}

\begin{algorithm}[t]
\caption{ReAct-Style Agent Loop}
\label{alg:react_stepwise}
\small
\begin{algorithmic}[1]
\Function{Agent}{$g, \mathcal{D}, \mathcal{E}, T$}
    \State $C_p \!\leftarrow\! \code{Retrieve}(\mathcal{D}, [g])$ \Comment{Retrieve for planning}
    \State $p \!\leftarrow\! \code[plan]{LLM}(g, C_p)$ \Comment{Generate high-level plan}
    \State $o_1 \!\leftarrow\! \mathcal{E}.\code{reset}(g)$
    \For{$t = 1$ to $T$}
        \State $C_t \!\leftarrow\! \code{Retrieve}(\mathcal{D}, [g, p, o_t])$ \Comment{Step-level retrieval}
        \State $r_t, a_t \!\leftarrow\! \code[react]{LLM}(g, p, o_t, C_t)$ \Comment{Reason and act}
        \State $o_{t+1}, \text{done} \!\leftarrow\! \mathcal{E}.\code{step}(a_t)$
        \If{done} \Return trajectory $\tau$ \EndIf
    \EndFor
    \Return trajectory $\tau$
\EndFunction
\end{algorithmic}
\end{algorithm}

Our approach collects teacher demonstrations once, then uses dynamic in-context learning and self-consistency cascades to route between student and teacher models at each agent decision step. Figure~\ref{fig:method} illustrates the overall logic of our algorithm.

\textbf{Demonstration collection phase:} We execute the teacher model $M_{\text{t}}$ on tasks from $\mathcal{T}_{\text{demo}}$ to collect a database $\mathcal{D}$ of complete trajectories, including observations, reasoning traces, and actions. This database is indexed using dense embeddings to enable fast similarity-based retrieval. Demonstration data collection is a one-time upfront cost; since we assume operating at scale with $|\mathcal{T}_{\text{test}}| \gg |\mathcal{T}_{\text{demo}}|$, this upfront cost is negligible when amortized over many test tasks. For readers interested in the exact crossover point where inference savings exceed demonstration costs, we provide a detailed analysis in Appendix~\ref{app:amortization}.

\textbf{Inference phase:} For each new task $(e,g) \in \mathcal{T}_{\text{test}}$, the student model $M_{\text{s}}$ executes a standard ReAct-style agent loop (Section~\ref{sec:background}). At each decision step $t$, the system retrieves the most relevant teacher examples from $\mathcal{D}$ based on the current goal, plan, and observation, and inserts them into the student's prompt (Section~\ref{sec:icd}). The student then samples multiple candidate actions under the same retrieved context. If all samples agree, the student's action is executed; if they diverge, the system defers to the teacher for that step (Section~\ref{sec:consistency}). This loop continues until the episode terminates, with total cost determined by the number of student queries (cheap) and teacher deferrals (expensive).

\subsection{Background: ReAct-Style Agent Architecture}
\label{sec:background}

The design of our agents follows the ReAct framework~\citep{yao2023react}, which interleaves planning, reasoning, and acting in a structured loop (Algorithm~\ref{alg:react_stepwise}). At the start of an episode, the agent generates a high-level plan $p$ for the goal $g$. At each timestep $t$, it observes the current state $o_t$, produces intermediate reasoning $r_t$, and selects an action $a_t$ to execute in the environment $\mathcal{E}$. Following recent best practices~\citep{kagaya2024rap,zhou2024trad,sarukkai:2025:selfgenerated}, we perform \emph{dynamic retrieval}—retrieving distinct exemplars for each step based on the current context rather than reusing a static prompt throughout the episode. We adopt this architecture as our foundation, building on recent findings that dynamic per-step retrieval improves multi-step agent performance.

The action $a_t$ at each step is obtained by querying a language model conditioned on the goal $g$, plan $p$, current observation $o_t$, and retrieved context $C_t$ from the teacher database. Our approach builds on this by using teacher-generated trajectories as the retrieval source, enabling cost reduction through a cheaper student model while maintaining the flexibility of this established architecture.

\subsection{Teacher Demonstration Database}
\label{sec:teacher_db}

We first construct a database $\mathcal{D} = \{\tau_i^{(\text{t})}\}_{i=1}^{|\mathcal{T}_{\text{demo}}|}$ of teacher trajectories using a high-performing model $M_{\text{t}}$ (e.g., \texttt{Claude Sonnet 4.5}). For each demonstration task $(e_i, g_i) \in \mathcal{T}_{\text{demo}}$, we run the agent using the teacher LLM in the environment $e_i$ to produce a complete trajectory:
\[
\tau_i^{(\text{t})} = \{g_i, p_i, (o_t^{(i)}, r_t^{(i)}, a_t^{(i)})_{t=1}^{T_i}\},
\]
where $p_i$ is the teacher's plan, and each step includes the observation $o_t$, reasoning trace $r_t$, and action $a_t$.

To enable retrieval, we compute dense embeddings using \texttt{MiniLM-L6-v2}~\citep{wang2020minilmv2}. For each trajectory, we separately embed the goal $\text{Embed}(g_i)$, plan $\text{Embed}(p_i)$, and at each step $t$, the reasoning $\text{Embed}(r_t^{(i)})$. These embeddings are indexed for fast similarity-based lookup during test-time deployment. Unlike prior self-improving agents that expand $\mathcal{D}$ online~\citep{sarukkai:2025:selfgenerated}, our setup keeps $\mathcal{D}$ fixed after initial collection to cleanly separate one-time teacher cost from ongoing student deployment cost.

\subsection{`Distillation' via Dynamic In-Context Learning}
\label{sec:icd}

At test time, a smaller frozen model $M_{\text{s}}$ (e.g., GPT-4.1-mini) interacts with new tasks $(e,g) \in \mathcal{T}_{\text{test}}$ using in-context exemplars retrieved from $\mathcal{D}$. The student reuses teacher reasoning traces as dynamic contextual guidance for each step, transferring behavior without weight updates.

\noindent\textbf{Multi-key retrieval.}
Inspired by the retrieval system in~\citet{sarukkai:2025:selfgenerated}, our retrieval operates on two levels:
\begin{itemize}[leftmargin=1.2em,itemsep=1pt]
    \item \textbf{Trajectory-level (planning):} Before execution, retrieve the $k$ most similar teacher goals based on cosine similarity. Extract the high-level plans $\{p_j\}_{j=1}^k$ from these trajectories and provide them as exemplars when prompting the student to generate its own plan $p$.
    \item \textbf{Step-level (acting):} At each step $t$, retrieve the top-$k$ most similar teacher steps using averaged cosine similarity across goal, plan, and reasoning. Return a window of neighboring steps $\{(o_j^*, r_j^*, a_j^*)\}$ from each retrieved trajectory—providing only the most relevant window of the trajectory.
\end{itemize}
This multi-key approach ensures retrieved examples match the current context across multiple relevant dimensions rather than relying on a single composite embedding.

These exemplars are inserted directly into the prompt of $M_{\text{s}}$:
\[
(r_t, a_t) \sim M_{\text{s}}\!\left(\cdot \mid g, p, o_t, \text{Retrieve}(\mathcal{D}; g, p, r_t)\right).
\]
Because retrieval occurs dynamically at each step, the student continually adapts to new contexts using only teacher demonstrations.

\noindent\textbf{Connection to reasoning distillation.}
Since retrieved trajectories contain both actions \emph{and} reasoning traces $r_t$ from the teacher, our approach can be viewed as an in-context variant of reasoning distillation~\citep{hsieh2023distilling}, which shows that distilling intermediate reasoning (not just final actions) improves student generalization. By including $r_t$ in the retrieved context, we enable the student to imitate not only \emph{what} the teacher did, but \emph{why}—providing richer signal for behavioral transfer.

Because distillation occurs via in-context learning rather than weight updates, it requires far fewer demonstration examples to effectively transfer knowledge. In our experiments we show this approach is effective using only a few hundred examples. This has important practical implications: (1) we can safely assume $|\mathcal{T}_{\text{demo}}| \ll |\mathcal{T}_{\text{test}}|$, making up-front demonstration costs negligible when amortized over large test sets, and (2) the cost of storing and retrieving from a database of a few hundred demonstrations is negligible. This synergy between sample-efficient in-context learning and cost-effective deployment at scale is a key advantage of our approach.

\subsection{Self-Consistency Based Deferral}
\label{sec:consistency}

While dynamic retrieval improves the student's ability to perform more tasks, there may remain situations where retrieved exemplars fail to effectively guide the student LLM. 
To detect when the student is likely to fail despite having access to retrieved teacher examples, we employ a lightweight introspection mechanism based on \emph{self-consistency}~\citep{wang2022selfconsistency}. The key insight is that when in-context learning succeeds, the student's understanding of the correct action should be stable across multiple samples. Conversely, when the retrieved examples provide insufficient guidance, there may be uncertainty in the student's output, manifesting as disagreement among samples.

\noindent\textbf{Why self-consistency rather than output uncertainty?}
LLM cascade methods~\citep{chen2023frugalgpt,schuster2022confident} sometimes use model confidence scores (e.g., token probabilities) to trigger deferral. In chain-of-thought settings, where the LLM first generates reasoning tokens before action tokens, computing the uncertainty in the generated answer either requires marginalizing out the reasoning path~\citep{wang2022selfconsistency}, or using a sampling-based approach to model the answer distribution~\citep{yue2024mot}, which naturally allows for the comparison of answers of different lengths. Paired with an LLM verifier, we find the sampling-based approach to be a simple and effective way to implement cascades in agentic settings with large action spaces.

\noindent\textbf{Implementation.}
At each step $t$, the student samples $N=3$ candidate actions conditioned on retrieved exemplars $C_t$:
\[
\{a_t^{(i)}\}_{i=1}^{N} \sim M_{\text{s}}(\cdot \mid g, p, o_t, C_t).
\]
By default, agreement requires all sampled actions to be identical:
\[
\text{Consistent}(\{a_t^{(i)}\}_{i=1}^{N}) = \mathbb{1}\!\left[\forall\, i,j,\; a_t^{(i)} = a_t^{(j)}\right].
\]
If consistent, we execute the student's action; otherwise, we defer to the teacher:
\[
a_t =
\begin{cases}
a_t^{(1)} & \text{if } \text{Consistent} = 1, \\
M_{\text{t}}(g, p, o_t) & \text{otherwise.}
\end{cases}
\]

For tasks with multiple valid formulations (e.g., API orchestration where different code snippets implement equivalent behavior), we use a \emph{soft-equivalence} variant: an auxiliary verifier assesses whether candidate actions are semantically equivalent despite syntactic differences.

Together, these mechanisms trigger teacher queries only when the student's in-context reasoning fails to produce consistent outputs.

\begin{algorithm}[t]
\caption{Dynamic In-Context Learning with Cascade}
\label{alg:cascade}
\small
\begin{algorithmic}[1]
\Function{Step}{$o_t, g, p, \mathcal{D}, M_\text{s}, M_\text{t}, N$}
    \State $C_t \!\leftarrow\! \text{Retrieve}(\mathcal{D}, [g,p,o_t])$ \Comment{Retrieve examples}
    \State $\{a_t^{(i)}\}_{i=1}^N \!\leftarrow\! \text{Sample}(M_\text{s}, g, p, o_t, C_t)$ \Comment{Sample student}
    \State $\text{consistent} \!\leftarrow\! \text{Consistent}(\{a_t^{(i)}\}_{i=1}^N)$ \Comment{Check agreement}
    \If{$\text{consistent}$}
        \State $a_t \!\leftarrow\! a_t^{(1)}$ \Comment{Use student action}
    \Else
        \State $a_t \!\leftarrow\! M_\text{t}(g, p, o_t)$ \Comment{Defer to teacher}
    \EndIf
    \State \Return $a_t$
\EndFunction
\end{algorithmic}
\end{algorithm}

\noindent\textbf{Cost accounting.}
A critical consideration is that sampling $N$ student outputs incurs $N$ times the student output token cost at each step. However, since $c_{\text{s}} \ll c_{\text{t}}$ (typically 10$\times$ cheaper), and costs are dominated by input tokens in the agentic setting, this overhead remains small compared to a single teacher query.

\section{Experiments}
\label{sec:experiments}

We structure our evaluation around three goals: 
(1) evaluate cost-accuracy trade-offs relative to baselines;
(2) isolate contributions of in-context learning vs. adaptive routing; and
(3) provide practitioners with guidance on key implementation choices.

\subsection{Benchmark selection}  
Since agents are commonly used to automate rote workflows—tasks that benefit from rapid prototyping and deployment—we chose benchmarks where task instances share structural patterns enabling cross-task learning, while exhibiting sufficient variety to require genuine adaptation. Please see Appendix~\ref{app:benchmark_details} for further details on benchmarks.

\textbf{AppWorld}~\citep{appworld-acl24} features workflow automation coding tasks requiring API composition (Gmail, Contacts, Calendar, etc.), user context interpretation, and multi-step state management. Tasks are evaluated via automated unit tests. We use 147 tasks from \texttt{train} and \texttt{val} as $\mathcal{T}_{\text{demo}}$, evaluating on 168 tasks in \texttt{test-normal}.

\textbf{ALFWorld}~\citep{shridhar2020alfworld} provides a controlled diagnostic on well-understood planning problems with discrete action spaces. We collect 500 teacher demonstrations from \texttt{train} as $\mathcal{T}_{\text{demo}}$ and evaluate on 134 tasks in \texttt{eval-out-of-distribution}.

We seek to reduce inference costs at scale, where $|\mathcal{T}_{\text{test}}| \gg |\mathcal{T}_{\text{demo}}|$. This mirrors deployment scenarios like batch processing where demonstration costs amortize over many instances. We only count inference costs on test sets, not demonstration collection costs, approximating execution at scale. Amortization analysis is in Appendix~\ref{app:amortization}.

\begin{figure*}[t]
    \centering
    \includegraphics[width=\textwidth]{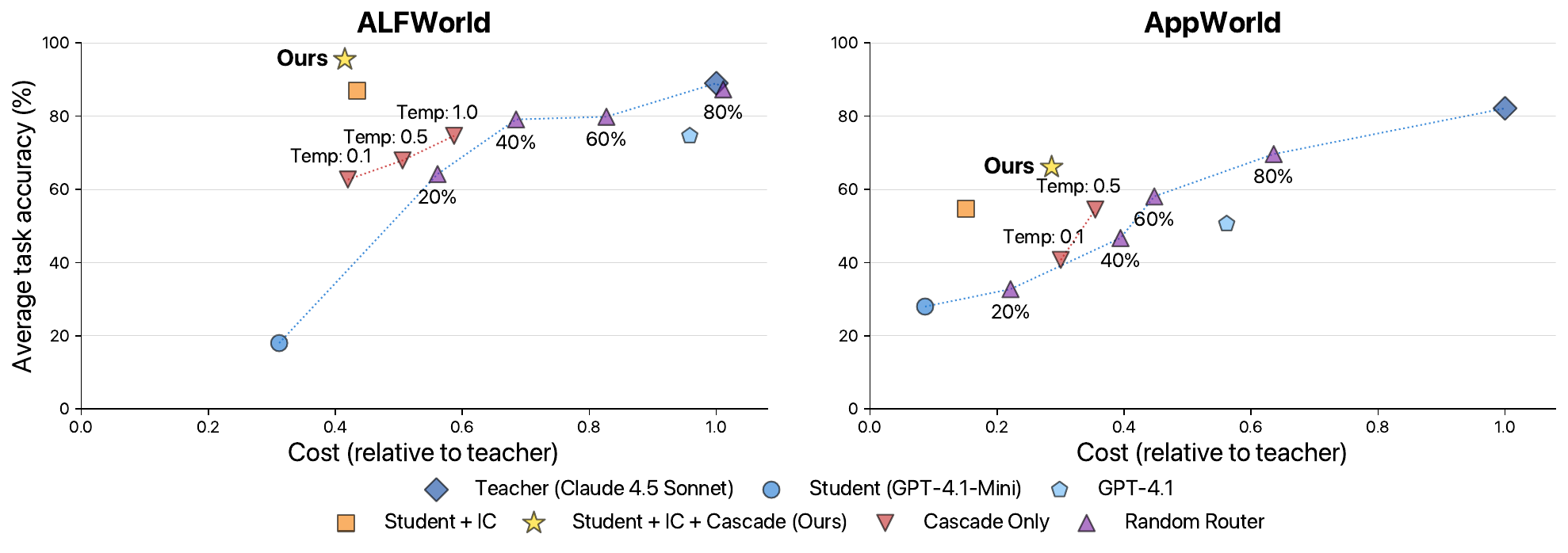}
    \vspace{-1em}
    \caption{\textbf{Combining in-context learning with cascades optimizes cost-accuracy tradeoffs.} Our IC + Cascade method breaks the Pareto frontier, performing better than the teacher on ALFWorld and significantly above others at similar cost on both benchmarks.}
    \vspace{-1em}
    \label{fig:pareto_overall}
\end{figure*}

\subsection{Teacher-student setup}
We use \texttt{Claude Sonnet 4.5} as teacher and \texttt{GPT-4.1-mini} as student. On AppWorld, \texttt{GPT-4.1-mini} serves as an LLM verifier for action equivalence. We validate generalization with \texttt{Llama-3.3-70B} as student to confirm our approach is not API-specific.

\emph{Core algorithmic configurations:}
\begin{itemize}[leftmargin=*,itemsep=0pt]
    \item \textbf{Teacher}: Upper bound on accuracy; ceiling cost.
    \item \textbf{Student (ZS)}: Zero-shot student; lower bound on cost, baseline accuracy. 
    \item \textbf{Student (IC)}: Student with retrieved teacher exemplars (tests dynamic in-context learning alone). 
    \item \textbf{Student (IC+Cascade)}: Full system (our primary method).
\end{itemize}

\emph{Alternative routing strategies:}
\begin{itemize}[leftmargin=*,itemsep=0pt]
    \item \textbf{Random Mix}: Query teacher for fixed fraction of steps (0.2, 0.4, 0.6, 0.8). Controls for whether routing value comes from adaptive gating vs. simply mixing teacher capacity.
    \item \textbf{Student (Cascade only)}: Self-consistency gating without exemplars, varying temperature (0.1, 0.5, 1.0). Isolates adaptive routing value without in-context improvement.
\end{itemize}

\emph{Ablations:}
\begin{itemize}[leftmargin=*,itemsep=0pt]
    \item \textbf{Retrieval strategies}: Compare retrieval approaches (number of examples $k$, retrieval keys, similarity metrics).
    \item \textbf{Demonstration set size}: Vary $|\mathcal{T}_{\text{demo}}|$ to characterize how many demonstrations are needed.
    \item \textbf{In-context mechanism analysis}: Investigate relationship between retrieved example content and student behavior replication.
\end{itemize}

\subsection{Metrics}
We measure task success rate as defined by each benchmark. For costs, we track all LLM calls: (i) student queries, (ii) teacher queries (when cascading), and (iii) LLM verifier calls (AppWorld only), recording input/output tokens and computing cost using current API pricing. We report costs normalized by Teacher baseline; see Appendix~\ref{app:implementation} for details.

\section{Results}

\subsection{Combining Dynamic In-Context Learning with Cascades Improves Cost-Accuracy Tradeoffs}
\label{sec:results:ic}

Figure~\ref{fig:pareto_overall} demonstrates that dynamic in-context learning from teacher examples and self-consistency cascades are complementary techniques that jointly optimize the cost-accuracy frontier.

\noindent\textbf{Dynamic in-context learning (alone) bridges the teacher-student gap.}
Figure~\ref{fig:pareto_overall} shows that Student (IC) establishes a new Pareto-optimal point between the zero-shot student and teacher baselines, outperforming Student (Cascade only) configurations at equivalent cost levels. On ALFWorld, Student (IC) achieves 0.87 accuracy—97\% of teacher accuracy (0.89) at 43\% of the cost (\$0.026 vs. \$0.059), compared to 0.18 zero-shot. On AppWorld, dynamic in-context learning improves accuracy from 0.28 to 0.55 while remaining at 15\% of teacher cost (\$0.089 vs. \$0.589).

\noindent\textbf{Combined system dominates alternatives.}
Our full algorithm, marked by the yellow stars on Figure~\ref{fig:pareto_overall}, establishes a new Pareto frontier on both benchmarks. On ALFWorld, we achieve 2.5$\times$ cost reduction at iso-quality compared to the teacher baseline, reducing inference costs from \$0.059 to \$0.024 per episode. At current API pricing, the upfront demonstration cost of \$29.50 amortizes after just 843 episodes, with cumulative savings exceeding \$34,900 for batch processing scenarios with 1M tasks (see Appendix~\ref{app:amortization} for full analysis). Notably, our combined system exceeds teacher accuracy (96\% vs. 89\%), likely because retrieved examples provide the student with both demonstration of the teacher's reasoning and implicit information about environment dynamics across multiple trajectories.

On AppWorld, our system achieves 3.5$\times$ cost reduction while recovering 79\% of teacher accuracy, demonstrating effective cost-performance trade-offs on this more challenging benchmark. Alternatively, when compared to the Pareto frontier obtained by randomly routing between student and teacher at different ratios, our system delivers a 0.28 boost in task success rate at iso-cost, while we achieve 2$\times$ cost reduction at iso-quality.

\begin{table}[h]
    \centering
    \begin{tabular}{lcc}
    \toprule
    \textbf{Method} & \textbf{GPT-4.1-mini} & \textbf{Llama-3.3-70B} \\
    \midrule
    \multicolumn{3}{l}{\textbf{ALFWorld}} \\[2pt]
    Teacher                 & 1.00 / 0.89 & 1.00 / 0.89 \\
    Student (ZS)            & 0.31 / 0.18 & 0.11 / 0.50 \\
    Student (IC)            & 0.43 / 0.87 & 0.14 / 0.87 \\
    Student (IC+Cascade)    & 0.41 / 0.96 & 0.21 / 0.93 \\[3pt]
    \midrule
    \multicolumn{3}{l}{\textbf{AppWorld}} \\[2pt]
    Teacher                 & 1.00 / 0.83 & 1.00 / 0.83 \\
    Student (ZS)            & 0.09 / 0.28 & 0.05 / 0.11 \\
    Student (IC)            & 0.15 / 0.55 & 0.10 / 0.32 \\
    Student (IC+Cascade)    & 0.29 / 0.66 & 0.19 / 0.44 \\
    \bottomrule
    \end{tabular}
    
    \caption{\textbf{Across multiple LLMs and benchmarks, Student (IC+Cascade) optimizes cost-accuracy tradeoffs.} Overall comparison of teacher–student configurations on ALFWorld and AppWorld.
    For both student LLMs, we use Sonnet 4.5 as the teacher. All numbers are formatted (Relative cost / Accuracy). 
    Cost is reported as fraction of teacher cost; Accuracy is episode success rate.}
    \label{tab:compare_models}

    \vspace{0em}

    \begin{tabular}{ccccc}
    \toprule
    \textbf{Difficulty} & \textbf{Teacher} & \textbf{\makecell{Student\\(IC+Casc.)}} & \textbf{\makecell{Student\\(ZS)}} & \textbf{Tasks} \\
    \midrule
    1 & 0.96 & 0.91 & 0.51 & 57 \\
    2 & 0.85 & 0.70 & 0.29 & 48 \\
    3 & 0.71 & 0.43 & 0.06 & 63 \\
    \bottomrule
    \end{tabular}
    \caption{\textbf{Student (IC+Cascade) is competitive with Teacher on easier AppWorld tasks, but struggles on harder tasks.} Task accuracy by difficulty level on AppWorld. Difficulty levels were manually assigned by benchmark authors based on planning and reasoning complexity~\citep{appworld-acl24}.}
    \label{tab:difficulty}
    \vspace{-1em}
\end{table}

\noindent\textbf{Our findings generalize to open-weight LLMs}
To test generality, we evaluate both closed- and open-weight models in structurally distinct domains. 
In Table~\ref{tab:compare_models}, trends remain consistent across sparse-reward embodied reasoning (ALFWorld) and multimodal app-use reasoning (AppWorld). 
On ALFWorld, Llama-3.3-70B reproduces the in-context gain (0.87 vs.\ 0.50 zero-shot) and exhibits comparable scaling under self-consistency deferral (boost to 0.93).
On AppWorld, Llama-3.3-70B is less effective than GPT-4.1-mini--but experiences similar benefits from in-context distillation and self-consistency deferral. 
These results indicate that exemplar reuse and confidence-based deferral generalize beyond proprietary APIs. 

\subsection{Contextualization: Comparing Against More Complex Alternatives}
\label{sec:results:alternatives}

Support for agile development workflows discourages reliance of fine-tuning and specialized system engineering. However to contextualize the cost-accuracy benefits of our approach we compare against these more complex alternatives.

\noindent\textbf{Our boosted GPT-4.1-mini offers better cost-accuracy tradeoffs than zero-shot GPT-4.1.} 
In Figure~\ref{fig:pareto_overall}, Zero-shot GPT-4.1 (marked with a green pentagon) achieves 51\% accuracy at 56\% relative cost on AppWorld—worse than our Student (IC) (55\% at 15\% cost) and far worse than Student (IC + Cascade) (66\% at 29\% cost). In-context distillation paired with intelligent deferral dominates fixed model selection. Similar trends hold for ALFWorld. Using a smaller LM (GPT-4.1-mini) with in-context examples from a frontier LM (Sonnet 4.5) can strictly dominate the use of a medium-cost LLM (GPT-4.1). 

\noindent\textbf{The accuracy of our simple approach is competitive with bespoke, state-of-the-art compound agentic systems.}
Our full method (Student IC + Cascade) achieves 65.5\% accuracy on AppWorld—approaching IBM's CuGA~\citep{marreed2025towards} (73.2\% on the leaderboard), a specialized compound system incorporating documentation retrieval, safety guardrails, multi-stage knowledge augmentation, and leverages task-specific modules and complex orchestration infrastructure. Our approach requires only teacher demonstrations, vector indexing, and self-consistency checking. In Section~\ref{sec:results:difficulty}, we show that if practitioners have access to task difficulty metadata, a simple difficulty-aware routing variant of our method reaches 77.6\% accuracy—surpassing CuGA. 

\noindent\textbf{Our approach is a training-free alternative to fine-tuning.} On ALFWorld, fine-tuning \texttt{GPT-4.1-mini} achieved 94\% accuracy (matching our 96\%). On AppWorld, we were unable to successfully fine-tune within our development timeline: API services failed when training on our teacher demonstrations, and open-weight models (\texttt{Llama-3.3-70B}) trained but produced 0\% task success despite converging on training loss. We see this outcome as evidence of the complexity of fine-tuning models for complex multi-step agents. It requires careful objective design and extensive debugging—challenges that are well-documented in prior work~\cite{barnett2024fine}. Our method provides a practical alternative for practitioners who seek rapid deployment without this overhead.

\subsection{Retrieving more in-context examples can boost task accuracy in exchange for higher costs}
\label{sec:results:retrieval}

\begin{figure}[t]
    \centering
    \begin{minipage}[t]{0.48\columnwidth}
        \centering
        \includegraphics[width=\textwidth]{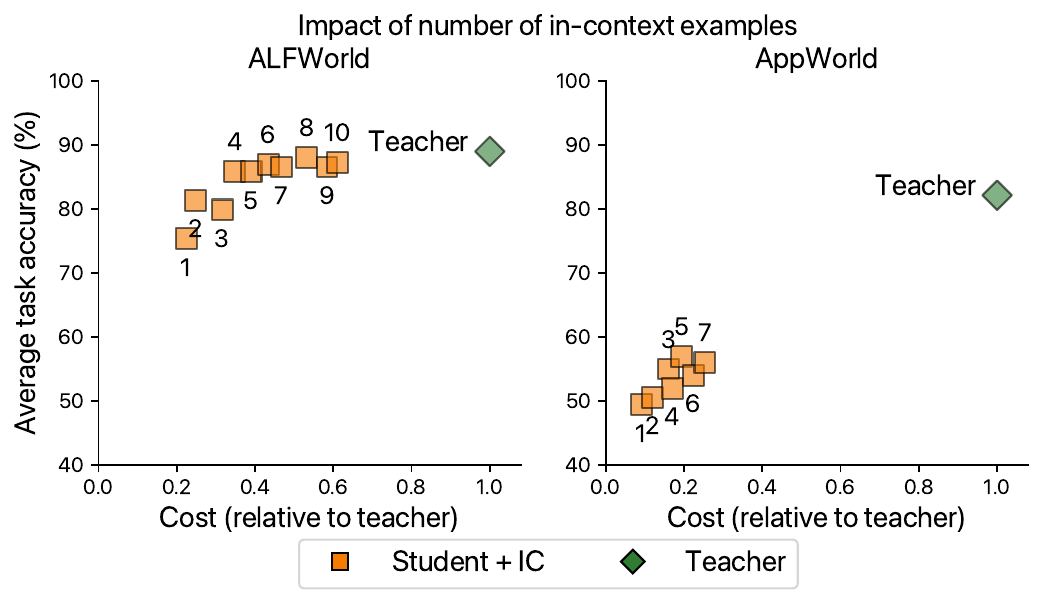}
        \caption{
        \textbf{Retrieving more in-context examples can boost task accuracy in exchange for higher costs}. Cost-accuracy tradeoff for varying numbers of retrieved in-context exemplars ($k$, labeled on each datapoint) on ALFWorld and AppWorld (Student IC, no cascade). On ALFWorld, accuracy improves rapidly from $k{=}1$ to $k{=}4$, then exhibits diminishing returns beyond $k{=}6$. On AppWorld, accuracy peaks at $k{=}5$ with more modest overall gains. In this paper, we use $k{=}6$ on ALFWorld and $k{=}3$ on AppWorld by default.
        } 
        \label{fig:ablate_ic}
    \end{minipage}
    \hfill
    \begin{minipage}[t]{0.48\columnwidth}
        \centering
        \includegraphics[width=\textwidth]{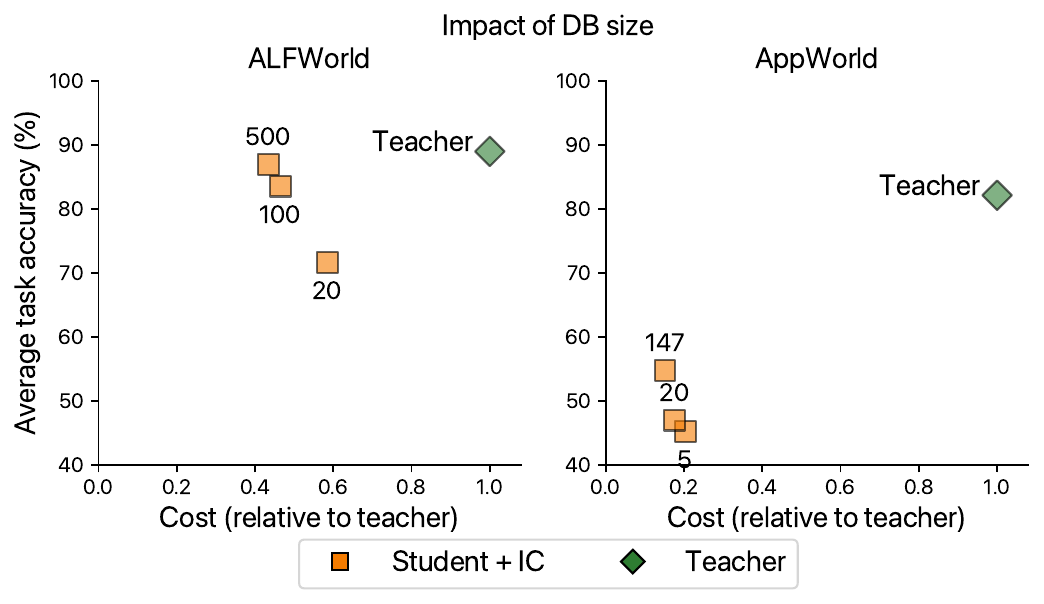}
        \caption{
        \textbf{Scaling teacher database size makes in-context distillation more effective.} The cost-accuracy tradeoff for varying teacher database sizes (labeled on each datapoint) on ALFWorld and AppWorld (Student+IC, no cascade). As we scale database size, more relevant examples are retrieved at each ReAct step, helping the agent solve tasks more successfully (increasing accuracy) and, as a corollary, more efficiently (reducing costs by shortening trajectories).
        }
        \label{fig:db_scaling}
    \end{minipage}
    \vspace{-1em}
\end{figure}

To guide practitioners in configuring in-context distillation systems, we vary the number of retrieved teacher exemplars ($k$) and measure the resulting cost-accuracy tradeoff on ALFWorld and AppWorld (Fig.~\ref{fig:ablate_ic}).

On ALFWorld, agent accuracy improves as more examples are added: accuracy rises from 0.75 at $k{=}1$ to 0.81 at $k{=}2$ (+7.8\%), and continues climbing to 0.86 at $k{=}4$ (+13.8\% from baseline). Beyond this point, improvements plateau---$k{=}6$ through $k{=}10$ yield accuracies between 0.86--0.88, representing only marginal gains (+2--3\%) despite costs increasing from 0.43$\times$ to 0.612$\times$ teacher cost. The cost-benefit analysis reveals diminishing returns: moving from $k{=}1$ to $k{=}4$ improves accuracy by 10.4 percentage points while increasing relative cost by 0.12 (54\% increase), whereas moving from $k{=}4$ to $k{=}10$ adds only 1.5 percentage points at a cost increase of 0.26 (76\% increase). In this paper, we chose $k=6$ on ALFWorld for our experiments. 

AppWorld exhibits a similar but more modest pattern. Accuracy improves from 0.49 at $k{=}1$ to a peak of 0.57 at $k{=}5$ (+7.6 percentage points), with costs rising from 0.09$\times$ to 0.19$\times$ teacher cost. However, the trend is less smooth---accuracy fluctuates between $k{=}3$ and $k{=}7$ (0.52--0.57), suggesting task-specific sensitivity to example selection. Beyond $k{=}5$, additional examples provide no consistent benefit while costs continue to scale linearly. In this paper, we chose $k=3$ on AppWorld for our experiments.

\subsection{Scaling teacher database size makes in-context distillation more effective.}
\label{sec:results:scaling}

A key practical question is how many teacher demonstrations are needed to achieve strong student accuracy. We vary the size of the teacher demonstration database and measure Student (IC) accuracy without cascading (Fig.~\ref{fig:db_scaling}).

In-context distillation exhibits strong data efficiency: on ALFWorld, just 100 teacher demonstrations yield 0.836 accuracy (94\% of teacher accuracy), while 500 demonstrations close the gap to 98\%. On AppWorld, 147 demonstrations (the full available training set) achieve 0.548 accuracy (67\% of teacher accuracy). Notably, even minimal databases (20 demonstrations on ALFWorld, 5 on AppWorld) substantially outperform zero-shot students (0.18 and 0.28 respectively), confirming that retrieval-based learning transfers effectively even from small exemplar sets.

Interestingly, larger databases also \emph{reduce cost} (AppWorld: 20\% of teacher costs at size 5 to 15\% at size 147, ALFWorld: 58\% of teacher costs at size 20 to 43\% at size 500). This occurs because better retrieval quality enables the agent to solve more tasks/take fewer steps to solve tasks, which in turn reduces total ReAct token usage. Our results suggest scaling teacher DB size may further close the accuracy gap, though we leave this to future work on AppWorld where dataset constraints currently limit exploration.

\subsection{Student failures are correlated with coverage gaps in retrieved examples}

We investigate the relationship between in-context examples and student behavior in our Student+IC setup (without Cascade) for AppWorld, hypothesizing that student failures occur primarily when retrieved examples fail to demonstrate needed behavior—\emph{coverage gaps}. We analyzed 53 tasks where the student (3-shot, no cascade) failed but the teacher succeeded. Using an LLM judge (Gemini-3.0-Flash), we found attribution in 48/53 cases (90.6\%), with 33 cases (68.8\%) specifically attributed to coverage gaps: retrieved examples did not demonstrate necessary API usage, disambiguation strategy, error handling, or other pivotal behavior. Remaining errors follow a long tail: contradictory examples (8.3\%), behavior shown but not applied (2.1\%), etc. Student performance is primarily governed by demonstration coverage—when retrieved examples lack necessary operations, students fail. This suggests demonstration collection should prioritize diversity of operations to ensure coverage. While LLM-based attribution is imperfect, we nevertheless believe this provides valuable insight. Future work could iteratively add demonstrations to fill identified coverage gaps. Prompts and qualitative examples are in Appendix~\ref{app:coverage-analysis}.

\subsection{Difficulty-Aware Deferral: Leveraging Domain Knowledge}
\label{sec:results:difficulty}

While self-consistency provides an effective automatic deferral signal, practitioners with domain knowledge about task difficulty can use it to inform routing decisions. We explore this using AppWorld's difficulty labels, where tasks are manually classified into three tiers based on planning complexity and API interaction requirements.

Table~\ref{tab:difficulty} shows that Student (IC + Cascade) accuracy degrades with difficulty but maintains substantial gains over Student (ZS) across all levels: 91\% vs 51\% on difficulty-1 (40pp gap), 70\% vs 29\% on difficulty-2 (41pp gap), and 43\% vs 6\% on difficulty-3 (37pp gap). This demonstrates the value of in-context learning across the difficulty spectrum, though the gap to Teacher widens on harder tasks (95\% → 82\% → 61\% recovery).

Practitioners with difficulty metadata can achieve different cost-accuracy tradeoffs. Using Student (IC + Cascade) for difficulty-1 and difficulty-2 while routing all difficulty-3 tasks to Teacher achieves 0.78 accuracy at 71\% of teacher cost—recovering 94\% of teacher accuracy with 29\% cost reduction. This represents a different operating point: compared to our automatic cascade (0.66 accuracy at 29\% cost), difficulty-based routing trades 2.4$\times$ higher cost for 12 percentage points of accuracy. When task difficulty signals are available and maximizing accuracy is critical, combining in-context distillation with domain-specific routing can shift the cost-accuracy tradeoff, though it sacrifices the deployment simplicity of purely automatic approaches.

\section{Related Work}
\label{sec:related}

\noindent\textbf{Knowledge distillation.}
Distillation trains a low-cost ``student'' to mimic a high-cost ``teacher''~\cite{Bucilu:2006:distill,hinton:2015:distillation}, optimizing student weights via gradient descent on teacher-labeled data. Recent trajectory-level distillation methods train smaller agents to imitate expert behavior~\citep{zeng2023agenttuning,chen2023fireact}. We also leverage teacher trajectories but perform distillation non-parametrically: trajectories serve as dynamically-retrieved in-context exemplars rather than supervised training targets. This eliminates fine-tuning costs and enables immediate deployment—the same frozen student applies to new tasks by adding demonstrations to the retrieval database.

\noindent\textbf{Dynamic in-context learning and retrieval.}
Modern LLMs adapt dramatically based on in-context demonstrations~\citep{yin2024deeper,agarwal2024many}. Prior retrieval-based approaches improve performance by dynamically selecting relevant examples~\cite{liu2021makes,su2022selective,qin2023ids}, but focus on single-query tasks. Recent work extends this to multi-step agent execution, using self-generated experience to improve agent capability over time~\citep{shinn2023reflexion,zhao2024expel,sarukkai:2025:selfgenerated}. We build on these approaches—particularly per-step retrieval mechanisms~\cite{sarukkai:2025:selfgenerated,zhou2024trad}—but apply them to cost reduction rather than capability improvement: we retrieve teacher-generated demonstrations at each step to guide a cheaper student model, enabling knowledge transfer without weight updates. Our contribution is operationalizing this for cost-efficient agentic workflows and characterizing how retrieval quality and demonstration coverage affect cost-accuracy tradeoffs. 

\noindent\textbf{Cascades and adaptive routing.}
Cascades improve cost-accuracy tradeoffs by routing easier instances to cheaper models and harder ones to expensive models~\cite{viola:jones:2001:cascade,wang:2022:wisdom,mamou:2022:tangobert,chen2023frugalgpt}. Prior work uses learned routers~\cite{li2021:cascade:bert,gupta:2024:languagecascade,chen2023frugalgpt,ong:2025:routellm} or model-intrinsic confidence scores~\cite{chow:1970:recogreject,Jitkrittum:2023:confidence,gupta:2024:languagecascade}, requiring training or model access. We adopt self-consistency-based routing~\cite{yue2024mot,wang2022selfconsistency}: sampling the student multiple times, where agreement signals confidence and disagreement triggers teacher deferral. This requires no learned components. Self-consistency has been used in single-model settings~\citep{manakul2023selfcheckgpt,wang2024softsc}; we apply it as a cascade routing signal. In our system, self-consistency evaluates the combined effectiveness of retrieval and in-context learning: relevant demonstrations yield convergent samples, while poor retrieval causes divergence.

\bibliography{mlsys_paper}
\bibliographystyle{iclr2026_conference}

\appendix

\section{Per-Step Retrieval Reduces Cost Without Sacrificing Accuracy}
\label{sec:results:retrieval_granularity}

A natural question is whether retrieving fresh examples at each step is necessary, or whether retrieving once at the start of the trajectory suffices. We compare our default \textbf{per-step retrieval}—which dynamically retrieves the top-$k$ most relevant trajectory windows at each decision point—against a \textbf{single retrieval} baseline that retrieves $k$ full trajectories once at the beginning and includes them in context for the entire episode.

Table~\ref{tab:retrieval_granularity} shows that per-step retrieval achieves equivalent accuracy while substantially reducing cost. On ALFWorld, both approaches achieve 87\% accuracy, but single retrieval incurs 54\% of teacher cost versus 43\% for per-step retrieval—a 26\% cost increase for no performance gain. On AppWorld, the pattern holds: accuracy remains essentially identical (55\% vs. 54\%), while the single-retrieval setup costs 60\% more (24\% vs. 15\% of teacher cost).

\begin{table}[h]
\centering
\caption{Comparison of retrieval granularity strategies. Per-step retrieval matches single-retrieval accuracy while reducing cost by dynamically selecting relevant windows rather than including full trajectories in context.}
\label{tab:retrieval_granularity}
\small
\begin{tabular}{lccc}
\toprule
\textbf{Method} & \textbf{Benchmark} & \textbf{Accuracy} & \textbf{Rel. Cost} \\
\midrule
Per-step retrieval & ALFWorld & 0.87 & 0.43 \\
Single retrieval & ALFWorld & 0.87 & 0.54 \\
\midrule
Per-step retrieval & AppWorld & 0.55 & 0.15 \\
Single retrieval & AppWorld & 0.54 & 0.24 \\
\bottomrule
\end{tabular}
\end{table}

The cost difference arises because single retrieval includes entire teacher trajectories (often 10-20 steps) in the student's context throughout execution, while per-step retrieval selects only the most relevant 3-5 step windows at each decision point. Since LLM API costs scale with input tokens, the longer context in single retrieval directly translates to higher costs.

Interestingly, the lack of accuracy difference suggests that \textbf{retrieval relevance matters more than context volume}. Including full trajectories does not help the student generalize better—it simply pays for irrelevant context. This validates our design choice to dynamically retrieve short, highly relevant windows rather than relying on static, comprehensive examples.

\section{Prompts}
\label{app:prompts}

We use identical prompt formats across both the ALFWorld and AppWorld benchmarks for all agent operations. The templates are deliberately minimal to avoid prompt engineering overhead: they specify the task structure and insert retrieved examples in the appropriate format (goals with plans for planning prompts, state-action-reasoning triples for execution prompts), but require no per-task customization.

Plan:
\begin{codebox}{python}
system_prompt: f'You are an expert at generating high-level plans of actions to achieve a goal.\n Here is your action space: {action_space}.\n Here are some examples of goal,plan from episodes that successfully achieved similar goals: {examples}'
user_prompt: f'goal: {goal}\n plan: '
\end{codebox}

ReAct:
\begin{codebox}{python}
system_prompt: f"You are a ReAct agent that is an expert at reasoning and taking actions to accomplish a goal. \n Goal: '{self.goal}'. \n You will receive observations and respond with *exactly* one reasoning and one action per observation. \n Write them on their own lines, with these exact labels and a colon: \n reasoning: <your step-by-step reasoning, concise, may be multi-line> \n action: <a single, directly executable command or final answer> \n Use the labels exactly as written: 'reasoning:' and 'action:' (no variations). \n Here is your action space: {action_space}.\n Here are some examples of goal, plan, observation, reasoning, action from episodes that successfully achieved similar goals: {examples}"
user_prompt: f'goal: {goal}\n plan: {plan}\n trajectory: {trajectory}\n action: '
\end{codebox}

Verifier:
\begin{codebox}{python}
system_prompt: f"You are an expert LLM judge that determines if a set of actions that could be taken by an LLM agent are all effectively equivalent. First, you will be provided the current state of the agent: the trajectory so far. You will receive a set of actions and respond with 'YES' if they are all equivalent and 'NO' if at least one is different. YOU MUST OUTPUT NOTHING ELSE. \n Here is your action space: {action_space}.\n Here are some examples of goal, plan, observation, reasoning, action from episodes that successfully achieved similar goals: {examples}"
user_prompt: f'goal: {goal}\n plan: {plan}\n trajectory: {trajectory}\n Set of actions: {set(individual_actions)}, effectively equivalent? '
\end{codebox}

\section{Implementation Details and Cost Model}
\label{app:implementation}

\subsection{Model Configuration}
Closed-weight models are queried via OpenAI/Anthropic APIs, with max 4096 output tokens for actions. All models are run with \texttt{temperature=0.1} unless otherwise noted. For our per-step retrieval mechanism (Section~\ref{sec:methods}), we use \texttt{MiniLM-L6-v2}~\citep{wang2020minilmv2} as the embedding model. All models use identical system prompts and task descriptions; only in-context exemplars vary. Full prompts in Appendix~\ref{app:prompts}.

\subsection{Token Usage and Cost Calculation Details}
\label{app:token_costs}

This appendix provides detailed token usage statistics and cost calculations for all methods evaluated in our experiments. All costs are computed by tracking every LLM call made during episode execution, including student queries, teacher queries (when cascading or randomly routing), and LLM verifier calls for semantic agreement checking (AppWorld only). While optimized KV caching of in-context examples has the potential to reduce costs further, we omit caching from our cost model and leave caching-aware costs as future work. We do not attribute any cost to the process of vector database lookup to retrieve examples. 

\subsection{Cost Calculation Methodology}

For each episode, we record input and output token counts for all model invocations and compute total cost using API pricing as of October 2025:

\begin{itemize}[leftmargin=*,itemsep=2pt]
    \item \textbf{GPT-4.1-mini}: \$0.40 per 1M input tokens, \$1.60 per 1M output tokens
    \item \textbf{GPT-4.1}: \$2.00 per 1M input tokens, \$8.00 per 1M output tokens
    \item \textbf{Claude Sonnet 4.5}: \$3.00 per 1M input tokens, \$15.00 per 1M output tokens
    \item \textbf{Llama-3.3-70B} (via NovitaAI, similar costs available on other providers): \$0.13 per 1M input tokens, \$0.39 per 1M output tokens
\end{itemize}

For each configuration, total episode cost is:
\begin{align*}
\text{Cost}_{\text{episode}} = \frac{1}{10^6} \bigg( &c_{\text{in}}^{(s)} \cdot T_{\text{in}}^{(s)} + c_{\text{out}}^{(s)} \cdot T_{\text{out}}^{(s)} \\
&+ c_{\text{in}}^{(t)} \cdot T_{\text{in}}^{(t)} + c_{\text{out}}^{(t)} \cdot T_{\text{out}}^{(t)} \bigg),
\end{align*}
where $T_{\text{in/out}}^{(s)}$ and $T_{\text{in/out}}^{(t)}$ denote total input/output tokens for student and teacher models respectively, and $c$ denotes the corresponding per-token prices.

Normalized cost is computed relative to the teacher baseline:
\[
\text{Cost}_{\text{normalized}} = \frac{\text{Cost}_{\text{episode}}}{\text{Cost}_{\text{teacher}}}.
\]

With Claude Sonnet 4.5 token pricing, the teacher baseline costs \$0.059 (USD) per episode on ALFWorld, and \$0.589 (USD) per episode on AppWorld. 

\subsubsection{ALFWorld Token Usage}

Table~\ref{tab:alfworld_tokens} reports average token usage and costs across 134 test episodes for all methods on ALFWorld. ``Blend'' denotes the random routing baseline at various fractions of teacher LLM calls (ex. Blend (0.2) is 20\% teacher LLM usage, 80\% student LLM usage).

\begin{table*}[h]
\centering
\caption{Detailed token usage and cost breakdown for ALFWorld. All values are averages per episode.}
\label{tab:alfworld_tokens}
\small
\begin{tabular}{lrrrrrrrr}
\toprule
\textbf{Method} & \makecell{\textbf{Student}\\\textbf{Input}} & \makecell{\textbf{Student}\\\textbf{Output}} & \makecell{\textbf{Teacher}\\\textbf{Input}} & \makecell{\textbf{Teacher}\\\textbf{Output}} & \makecell{\textbf{Steps}\\\textbf{per Task}} & \makecell{\textbf{Teacher}\\\textbf{Frac}} & \textbf{Acc} & \makecell{\textbf{Cost}\\\textbf{(norm.)}} \\
\midrule
Teacher & — & — & 16257 & 684 & 12 & — & 0.89 & 1.0 \\
\midrule
Student (ZS) & 41457 & 1130 & — & — & 27 & 0.00 & 0.18 & 0.31 \\
Student (IC) & 61966 & 527 & — & — & 12 & 0.00 & 0.87 & 0.43 \\
Student (IC + Cascade) & 50648 & 1283 & 597 & 26 & 9.9 & 0.039 & 0.96 & 0.42 \\
Student (Cascade only) & 26669 & 2194 & 2910 & 125 & 19 & 0.10 & 0.63 & 0.42 \\
\midrule
Blend (0.2) & 22561 & 683 & 6496 & 234 & 20 & — & 0.64 & 0.56 \\
Blend (0.4) & 13183 & 436 & 9657 & 365 & 16 & — & 0.79 & 0.68 \\
Blend (0.6) & 7813 & 276 & 12670 & 482 & 14 & — & 0.80 & 0.83 \\
Blend (0.8) & 3106 & 149 & 16316 & 616 & 13 & — & 0.87 & 1.0 \\
\midrule
GPT-4.1 (ZS) & 25400 & 720 & — & — & — & 0.00 & 0.75 & 0.96 \\
\midrule
Llama-3.3-70B (ZS) & 44655 & 2275 & — & — & — & 0.00 & 0.50 & 0.11 \\
Llama-3.3-70B (IC) & 61727 & 951 & — & — & — & 0.00 & 0.87 & 0.14 \\
Llama-3.3-70B (IC + Cascade) & 55706 & 2385 & 1113 & 50 & — & 0.06 & 0.93 & 0.21 \\
\bottomrule
\end{tabular}
\end{table*}

\subsubsection{AppWorld Token Usage}

Table~\ref{tab:appworld_tokens} reports average token usage and costs across 168 test episodes for all methods on AppWorld. ``Blend'' denotes the random routing baseline at various fractions of teacher LLM calls (ex. Blend (0.2) is 20\% teacher LLM usage, 80\% student LLM usage).

\begin{table*}[h]
\centering
\caption{Detailed token usage and cost breakdown for AppWorld. All values are averages per episode.}
\label{tab:appworld_tokens}
\small
\begin{tabular}{lrrrrrrrr}
\toprule
\textbf{Method} & \makecell{\textbf{Student}\\\textbf{Input}} & \makecell{\textbf{Student}\\\textbf{Output}} & \makecell{\textbf{Teacher}\\\textbf{Input}} & \makecell{\textbf{Teacher}\\\textbf{Output}} & \makecell{\textbf{Steps}\\\textbf{per Task}} & \makecell{\textbf{Teacher}\\\textbf{Frac}} & \textbf{Acc} & \makecell{\textbf{Cost}\\\textbf{(norm.)}} \\
\midrule
Teacher & — & — & 185460 & 2183 & 20 & — & 0.82 & 1.0 \\
\midrule
Student (ZS) & 118430 & 2317 & — & — & 19 & 0.00 & 0.28 & 0.087 \\
Student (IC) & 216563 & 1623 & — & — & 12 & 0.00 & 0.55 & 0.15 \\
Student (IC + Cascade) & 251683 & 4829 & 18385 & 329 & 12 & 0.22 & 0.66 & 0.29 \\
Student (Cascade only) & 184487 & 7009 & 28445 & 440 & 18 & 0.21 & 0.41 & 0.30 \\
\midrule
Blend (0.2) & 96831 & 1809 & 27597 & 398 & 18 & — & 0.33 & 0.22 \\
Blend (0.4) & 85588 & 1520 & 61349 & 778 & 19 & — & 0.47 & 0.39 \\
Blend (0.6) & 57568 & 1018 & 74379 & 1066 & 18 & — & 0.58 & 0.45 \\
Blend (0.8) & 23362 & 376 & 113894 & 1527 & 17 & — & 0.69 & 0.64 \\
\midrule
GPT-4.1 (ZS) & 155035 & 2603 & — & — & — & 0.00 & 0.51 & 0.56 \\
\midrule
Llama-3.3-70B (ZS) & 198953 & 4893 & — & — & — & 0.00 & 0.11 & 0.047 \\
Llama-3.3-70B (IC) & 445542 & 3438 & — & — & — & 0.00 & 0.32 & 0.10 \\
Llama-3.3-70B (IC + Cascade) & 324814 & 6616 & 20749 & 419 & — & 0.13 & 0.44 & 0.19 \\
\bottomrule
\end{tabular}
\end{table*}

\subsubsection{Key Observations}

\paragraph{Token efficiency of dynamic in-context learning.}
Comparing Student (ZS) to Student (IC), we observe that adding retrieved exemplars \emph{reduces} trajectory length (ALFWorld: 27.0 → 12.1 steps; AppWorld: 19.3 → 12.1 steps) despite increasing input tokens per step. This occurs because better-guided students solve tasks more directly, avoiding exploratory dead ends. The net effect is higher input token cost per step but fewer steps overall, yielding favorable cost-performance tradeoffs.

\paragraph{Cascade overhead.}
Self-consistency cascades add sampling overhead (multiple student queries per step) but trigger teacher queries on only a small fraction of steps (ALFWorld: 3.9\%; AppWorld: 22.2\%). Since student queries cost 10-100$\times$ less than teacher queries, the net cost increase from cascading is modest (ALFWorld: 0.434 → 0.415 normalized cost; AppWorld: 0.151 → 0.286).

\paragraph{Teacher fraction vs. cost.}
Comparing Blend configurations shows that cost scales approximately linearly with teacher fraction, but random routing provides much worse accuracy than adaptive cascades at equivalent teacher usage (e.g., AppWorld Blend 0.2 vs. IC + Cascade: similar teacher fraction but 0.221 vs. 0.286 normalized cost due to different student base capability).

\subsubsection{Amortization Analysis: When Do Inference Savings Offset Demonstration Costs?}
\label{app:amortization}

A practical concern for deployment is whether the upfront cost of collecting teacher demonstrations is justified by inference-time savings. We analyze the breakeven point where cumulative inference cost reductions exceed the one-time cost of constructing the teacher database $\mathcal{D}$.

\paragraph{Setup.}
Let $C_{\text{demo}}$ denote the total cost of collecting teacher demonstrations on $\mathcal{G}_{\text{train}}$, computed as:
\[
C_{\text{demo}} = |\mathcal{G}_{\text{train}}| \cdot C_{\text{teacher}}^{\text{episode}},
\]
where $C_{\text{teacher}}^{\text{episode}}$ is the average per-episode cost of running the teacher model (reported in Tables~\ref{tab:alfworld_tokens} and~\ref{tab:appworld_tokens}).

At deployment, let $C_{\text{baseline}}$ denote the per-episode cost of the baseline approach (e.g., always using the teacher), and $C_{\text{ours}}$ denote the per-episode cost of our Student (IC + Cascade) method. After processing $N$ test episodes, the total cost including demonstration collection is:
\begin{align*}
\text{Total}_{\text{baseline}} &= N \cdot C_{\text{baseline}}, \\
\text{Total}_{\text{ours}} &= C_{\text{demo}} + N \cdot C_{\text{ours}}.
\end{align*}

The breakeven point $N^*$ occurs when $\text{Total}_{\text{ours}} = \text{Total}_{\text{baseline}}$:
\[
N^* = \frac{C_{\text{demo}}}{C_{\text{baseline}} - C_{\text{ours}}}.
\]

\paragraph{ALFWorld breakeven analysis.}
Using costs from Table~\ref{tab:alfworld_tokens}:
\begin{itemize}[leftmargin=*,itemsep=2pt]
    \item Teacher demonstration cost: $C_{\text{demo}} = 500 \times \$0.059 = \$29.50$
    \item Baseline (Teacher) per episode: $C_{\text{baseline}} = \$0.059$
    \item Student (IC + Cascade) per episode: $C_{\text{ours}} = \$0.024$
    \item Per-episode savings: $\Delta C = \$0.059 - \$0.024 = \$0.035$
\end{itemize}

Breakeven point:
\[
N^* = \frac{\$29.50}{\$0.035} \approx 843 \text{ episodes}.
\]

After processing 843 test episodes, cumulative savings offset the upfront demonstration cost. For a batch processing scenarios with 1,000,000 tasks, over \$34,900 is saved.

\paragraph{AppWorld breakeven analysis.}
Using costs from Table~\ref{tab:appworld_tokens}:
\begin{itemize}[leftmargin=*,itemsep=2pt]
    \item Teacher demonstration cost: $C_{\text{demo}} = 147 \times \$0.59 = \$86.73$
    \item Baseline (Teacher) per episode: $C_{\text{baseline}} = \$0.59$
    \item Student (IC + Cascade) per episode: $C_{\text{ours}} = \$0.17$
    \item Per-episode savings: $\Delta C = \$0.59 - \$0.17 = \$0.42$
\end{itemize}

Breakeven point:
\[
N^* = \frac{\$86.73}{\$0.42} \approx 207 \text{ episodes}.
\]

After processing just 207 test episodes, the method becomes cost-effective. For deployment on 1,000,000 tasks, over \$419,000 is saved.

\subsection{Retrieval Costs}
We do not attribute any cost to the process of vector database lookup to retrieve examples. Since our databases include only a few hundred demonstrations, it is feasible to make these costs negligible in practice using standard vector search infrastructure.

\section{Benchmark Details}
\label{app:benchmark_details}

\subsection{Real-World Motivation}
One of the most common use of agents in real-world industry practice involves automating rote workflows such as common business processes, processing customer service tickets, and implementing data processing pipelines. These tasks benefit from rapid prototyping and deployment without extensive model training.

\subsection{AppWorld}
AppWorld tasks mirror realistic daily business workflows. Tasks are evaluated via automated unit tests checking both state changes and execution traces, ensuring that agents not only achieve the correct final state but also follow reasonable execution patterns.

\subsection{ALFWorld}
ALFWorld is a standard benchmark in several prior works on agentic self-improvement~\citep{zhao2024expel,fu2024autoguide,chen2024automanual,sarukkai:2025:selfgenerated}. The availability of a large train set (up to 3500 tasks) enables us to test how our method scales with varying demonstration set sizes.

\subsection{Scale and Amortization}
Few academic benchmarks (including AppWorld and ALFWorld) exhibit the scale where $|\mathcal{T}_{\text{test}}| \gg |\mathcal{T}_{\text{demo}}|$ by orders of magnitude. In practice, deployment scenarios like large-scale batch processing or production agent services would process many more test instances, making demonstration collection costs increasingly negligible. Our experimental setup approximates this by excluding demonstration costs from reported metrics.

\section{Mechanistic Role of In-Context Examples in Student Behavior}
\label{app:coverage-analysis}

On AppWorld, to understand why the student model with in-context examples (3ic) failed on tasks where the teacher model succeeded, we performed a two-stage analysis. First, we identified the ``difference maker'' step--the point where, faced with similar states, the teacher took a fundamentally different action that led to success while the student's approach led to failure. Second, we attributed the student's behavior at this pivotal step to the content of the in-context examples provided, examining whether the examples demonstrated the necessary behavior, provided conflicting guidance, or failed to teach critical patterns. Both analyses used the Gemini-3.0-Flash. The main results of the analysis are included in the main paper--we include both an anecdotal example and the prompts used to conduct the analysis below.

\subsection{Anecdotal example of insufficient coverage in retrieved in-context examples}

\textbf{Task:} AppWorld task 6 (test set) requires removing a set of songs from a Spotify queue.

\textbf{Teacher behavior:} The teacher model successfully completes the task by deleting songs back-to-front from the queue, using a specific deletion strategy that maintains queue indices correctly.

\textbf{Student behavior (3-shot):} The student with in-context examples fails to remove the songs successfully.

\textbf{Retrieved in-context examples:} The system retrieved three teacher demonstrations (training examples 47, 55, 58). Example 55 demonstrates how to retrieve the song queue, but critically, \emph{none of the three demonstrations show an effective strategy to delete songs from the queue}. Without an example demonstrating the deletion operation, the student lacks the necessary guidance to execute this pivotal step, leading to task failure.

\textbf{Analysis:} This example illustrates a clear coverage gap—the operation needed (queue deletion) was not demonstrated in any of the retrieved examples. The student had examples of related operations (queue retrieval) but lacked coverage of the specific operation required for success.

\subsection{Prompt for identifying divergent step between student and teacher}

This prompt identifies the ``difference maker'' step where the teacher succeeded but the student failed.

\begin{codebox}{python}
f"""You are analyzing two trajectories for the same task. One trajectory is 
from a student model (student) which FAILED, and one is from a teacher 
model (teacher) which SUCCEEDED.

Your task is to identify the "difference maker" step - the point where, 
faced with the same or similar state/situation, the teacher took a 
fundamentally different action or strategy that led to success, while 
the student's approach led to failure.

IMPORTANT CONSIDERATIONS:
1. The step numbers in each trajectory may be DIFFERENT (e.g., step 5 
   in student vs step 8 in teacher)
2. Look for points where both trajectories were in similar states 
   (similar observations, similar context)
3. Identify where one trajectory made a strategic choice that the 
   other didn't
4. This may NOT be the first difference - trajectories might converge 
   and diverge multiple times
5. Focus on the decision that represents the most significant strategic 
   difference

Consider a "difference maker" if:
- The teacher used a different API or approach that solved a critical 
  problem the student missed
- The teacher recognized a prerequisite or dependency the student 
  overlooked
- The teacher handled an error or edge case that caused the student 
  to fail
- The teacher took a more direct or correct path to the solution
- The teacher demonstrated better understanding of the problem domain 
  or API requirements

Do NOT consider it a "difference maker" if:
- It's just a minor variation in code syntax
- It's the same strategy with different variable names
- The trajectories were already on different paths due to earlier 
  differences

Here are the steps from both trajectories:

STUDENT TRAJECTORY (3ic):
[Steps with reasoning, action, and observation for each step]

TEACHER TRAJECTORY (zeroic):
[Steps with reasoning, action, and observation for each step]

Please respond with ONLY a JSON object in this exact format:
{
  "student_step": <step_number_in_student_trajectory>,
  "teacher_step": <step_number_in_teacher_trajectory>,
  "student_action": "<the action taken in student at that step>",
  "teacher_action": "<the action taken in teacher at that step>",
  "similar_state_description": "<description of the similar state/
    situation both trajectories were in>",
  "explanation": "<detailed explanation of why the teacher's approach 
    succeeded where the student's failed>",
  "why_teacher_succeeded": "<explanation of what the teacher did 
    correctly that the student missed>",
  "key_lesson": "<the key lesson or insight that the student should 
    learn from the teacher's approach>"
}

If you cannot find a clear difference maker, respond with:
{
  "student_step": null,
  "teacher_step": null,
  "explanation": "<explanation of why no clear difference maker was 
    found>"
}"""
\end{codebox}

\subsection{Prompt for checking relationship between student action and in-context examples}

This prompt analyzes whether pivotal student behavior should be attributed to in-context examples.

\begin{codebox}{python}
f"""You are analyzing how in-context examples influenced a model's 
behavior at a pivotal step.

The student model WITH in-context examples (3ic) failed, while the 
teacher model succeeded.

PREVIOUS ANALYSIS RESULT:
[The JSON output from the difference maker step analysis]

PIVOTAL STEP INFORMATION:
- Step number in student trajectory: {pivotal_step}
- Student file: {student_file}

What content from the in-context examples may have led the 3ic model 
to take the failing action, or what did the examples fail to teach 
that the teacher model knew?

IN-CONTEXT EXAMPLES PROVIDED AT THIS STEP:
The following examples were provided to the student model at step 
{pivotal_step}:
[Example content snippets]

Please analyze:
1. Is there content in these examples that directly shows or suggests 
   the successful/failing action?
2. Are there patterns, strategies, or API usage patterns in the 
   examples that may have influenced the behavior?
3. For the failing case: Did the examples teach something incorrect, 
   or fail to teach something important?
4. For the succeeding case: What specific content from the examples 
   led to the correct approach?

Please respond with ONLY a JSON object in this exact format:
{
  "attribution_found": <true_or_false>,
  "example_ids_involved": [<list_of_example_ids_that_may_have_
    influenced>],
  "specific_content": "<specific text or patterns from examples that 
    may have influenced the behavior>",
  "how_it_influenced": "<explanation of how the example content 
    influenced the behavior>",
  "confidence": "<high/medium/low - confidence in this attribution>",
  "alternative_explanation": "<if attribution is low confidence, what 
    else might explain the behavior?>"
}

If no clear attribution can be made, set "attribution_found" to false 
and explain why."""
\end{codebox}

\section{Teacher Performance with In-Context Examples}
\label{app:teacher_ic}

In Section~\ref{sec:results:ic}, we report that our Student (IC + Cascade) method achieves 94\% accuracy on ALFWorld, exceeding the baseline Teacher accuracy of 89\%. A natural question is whether this performance gain comes from the cascade mechanism selecting high-quality teacher responses, or from the boost provided by in-context examples themselves.

To contextualize this result, we ran an additional experiment: Teacher + IC, where the teacher model (\texttt{Claude Sonnet 4.5}) is provided with the same retrieved in-context examples at each step as the student receives. This configuration achieves \textbf{94\% accuracy}, matching Student (IC + Cascade).

This result confirms that the performance gain is attributable to in-context learning: when relevant teacher demonstrations are provided as examples, both teacher and student models benefit. The Student (IC + Cascade) configuration matches teacher performance by combining two mechanisms: (1) boosting the student with in-context examples when they provide consistent guidance, and (2) cascading to the (boosted) teacher when student samples diverge. The fact that both Teacher + IC and Student (IC + Cascade) reach 94\% suggests that in-context learning lifts the performance ceiling on this benchmark, even for a teacher learning from its own examples, and our cascade mechanism successfully navigates between the boosted student and teacher to achieve this ceiling cost-efficiently.

\end{document}